\documentclass[jounal]{IEEEtran}

\usepackage{amsmath}
\usepackage{amssymb,amsthm}
\usepackage{graphicx}
\usepackage{psfrag}
\usepackage{epsfig}
\usepackage{dblfloatfix}
\usepackage{subfigure}
\usepackage{color}
\usepackage{latexsym, amssymb, array, multirow}
\usepackage{algorithm, algpseudocode}

\makeatletter
\def\BState{\State\hskip-\ALG@thistlm}
\makeatother

\newtheorem{proposition}{Proposition}
\theoremstyle{definition}

\theoremstyle{remark}

\def\tr{\mathop{\rm tr}\nolimits}%
\def\diag{\mathop{\rm diag}\nolimits}%
\def\Re{\mathop{\rm Re}\nolimits}%
\def\Im{\mathop{\rm Im}\nolimits}%

\newcommand{\sr}{{\rm SR}}
\newcommand{\sd}{{\rm SD}}
\newcommand{\rd}{{\rm RD}}

\newcommand{\rt}{{\rm R}}
\newcommand{\dt}{{\rm D}}
\newcommand{\zf}{{\rm ZF}}
\newcommand{\sign}{{\rm sign}}
\newcommand{\relu}{{\rm Relu}}
\newcommand{\xv}{{\bf x}}
\newcommand{\yv}{{\bf y}}
\newcommand{\zv}{{\bf z}}
\newcommand{\vv}{{\bf v}}
\newcommand{\uv}{{\bf u}}
\newcommand{\Wv}{{\bf W}}
\newcommand{\bv}{{\bf b}}
\newcommand{\iv}{{\bf i}}

\newcommand{\tv}{{\bf t}}
\newcommand{\Xv}{{\bf X}}
\newcommand{\Lv}{{\bf L}}

\newcommand{\sv}{{\bf s}}

\newcommand{\CN}{\mathcal{CN}}
\newcommand{\N}{\mathcal{N}}
\newcommand{\Complex}{\mathbb{C}}

\begin{document}

\title{Deep Learning Detection Networks in MIMO Decode-Forward Relay Channels}

\author{Xianglan~Jin,~\IEEEmembership{Member,~IEEE,}
        and~Hyoung-Nam Kim,~\IEEEmembership{Member,~IEEE}
\thanks{X. Jin and H.-N. Kim are with the Department of Electronics
Engineering, Pusan National University, Busan 46241, Republic of Korea, (e-mails: jinxl77@gmail.com, hnkim@pusan.ac.kr,
Phone number: +82-51-510-2394.) } 
\thanks{This research has been supported by Basic Science Research Program through the National Research Foundation of Korea (NRF) funded by the Ministry of Education under Grant NRF-2017R1D1A1A09000565. }}

\maketitle

\begin{abstract}
In this paper, we consider signal  detection algorithms in a multiple-input multiple-output (MIMO) decode-forward (DF) relay channel with one source, one relay, and one destination.
The existing suboptimal near maximum likelihood (NML) detector and the NML with two-level pair-wise error probability  (NMLw2PEP) detector achieve excellent performance with instantaneous channel state information (CSI) of the source-relay (SR) link  and with statistical CSI of the SR link, respectively.
However, the NML detectors require an exponentially increasing complexity as the number of transmit antennas increases.
Using deep learning algorithms,  NML-based detection networks (NMLDNs) are proposed with and without the CSI of the SR link at the destination.
The NMLDNs detect signals in changing channels after a single training using a large number of randomly distributed channels.
The detection networks require much lower detection complexity than the exhaustive search NML detectors while exhibiting good performance.
To evaluate the performance, we introduce  semidefinite relaxation  detectors with polynomial complexity based on the NML detectors.
Additionally, new linear detectors based on the zero gradient of the NML metrics are proposed.
Applying various detection algorithms at the relay (DetR) and detection algorithms at the destination (DetD), we present  some DetR-DetD methods in MIMO DF relay channels.
An appropriate  DetR-DetD method can be employed according to the required error probability and detection complexity.
The complexity analysis and simulation results validate the arguments of this paper.
\end{abstract}
\begin{IEEEkeywords}
 Channel state information, machine learning, maximum likelihood,  neural network, TensorFlow.
\end{IEEEkeywords}

\IEEEpeerreviewmaketitle
\section{Introduction}
In wireless communications, deep fading often causes a failure in reliable data transmission. 
Relays help increase the transmission reliability between a source and a destination, and extend the network coverage by providing an additional link. 
The relay channel model, introduced by van der Meulen \cite{van-der-Meulen1971b}, is a basic channel model in network communications.
This relay channel has been studied extensively in the literature
\cite{Cover--El-Gamal1979}-\cite{Jin--Kim2017}.
Among various relaying operations, amplify-forward (AF) and decode-forward (DF) are the two most common methods \cite{Laneman--Tse--Wornell2004}.
 
Unlike a multiple-input multiple-output (MIMO) system, a linear relationship does not exist between the input and the output in the DF relay channel due to the hard decision at the relay.
As the received signal of the relay is not known at the destination, the maximum likelihood (ML) detection in the DF relay system requires more steps than that in the MIMO system \cite{Sendonaris2}, \cite{Wang--Cano--Giannaki--Laneman2007}, \cite{Jin2011ieice}.
Due to the complexity of the ML detection and the difficulty in analysis,  a near-ML (NML) detector was proposed in \cite{Jin2011ieice}, \cite{Jin2011twc} under instantaneous channel state informations (CSIs) of the source-relay (SR), source-destination (SD), and relay-destination (RD) links.
However, forwarding the instantaneous CSI of the SR link from the relay to the destination requires additional work and reduces the data rate.
With the statistical CSI of the SR link at the destination, an NML with two-level pair-wise error probability (PEP) (NMLw2PEP) detection was proposed in \cite{Jin2014jcn}  that achieves good performance with relatively low complexity.
Without any knowledge of the SR link at the destination,  the minimum distance (MD) detection\footnote{This was called  a maximum ratio combining (MRC) in a single-antenna relay system in \cite{Wang--Cano--Giannaki--Laneman2007}.} ignores detection error at the relay and shows very poor performance  \cite{Wang--Cano--Giannaki--Laneman2007}, \cite{Jin2011ieice}.
The above mentioned detection algorithms detect signals simultaneously by exhaustively searching all the possible signal sets so that their complexities increase exponentially as the number of transmit antennas increases.

A method to reduce the detection complexity is to separate the signals by a linear operation and detect them individually.
The typical linear detectors in MIMO channels are the zero forcing (ZF) and minimum mean square estimation (MMSE) detectors \cite{Lupas--Verdu1989}.
Referring to the ZF detector,  a linear detector of ZF with maximum ratio combining (MRC) (ZFwMRC) was proposed in  MIMO relay channels when the relay detects signals correctly \cite{Chalise--Vandendorpe2009}.
However, this algorithm cannot achieve good performance for the relay with errors similar to the MD detection.
A new detection method should be introduced, and a potential solution is to use the powerful tools in machine learning.

\subsection{Machine Learning and Detection}
Machine learning is a subset of artificial intelligence that learns to solve a specific problem by themselves \cite{Goodfellow--Bengio--Courville2016}. 
Supervised learning, a basic machine learning algorithm,  trains a learning algorithm, $g$, which is an approximate of a target function $f$ such that $x=f(y)$ using the known training data samples  including  the observation data $y$ and reference data $x$.
Meanwhile, traditional signal detection obtains an estimation of $\hat{x}$ directly from the observation $y$ using a mathematical optimization method without reference signals (training data).
However, it is not easy to theoretically find a detector with reasonable performance and complexity.
Applying machine learning, a learning algorithm, $g$, that approximates the existing detection algorithm is trained to minimize a loss function $l(\hat{x},x)$ that measures the cost of estimating $\hat{x}$ when the actual answer is $x$.
After training, the observation data $y$ undergoes the final learning algorithm $g$; and subsequently, the desired data $\hat{x}$ is detected in real time. 
This testing phase is called a detection stage in this paper.

Advances in computer technology and big data processing have significantly reduced the cost and time of training deep learning algorithms.
This has significantly improved the development of computer vision \cite{Ioannidou--Chatzilari--Nikolopoulos--Kompatsiaris} and natural language processing \cite{Abdel-Hamid--Mohamed--Jiang--Deng--Penn--Yu2014}.
In communication networks, deep learning has begun to receive much attention \cite{Zhang--Patras--Haddadi2018}.
To reduce complexity, the detection and channel decoding problems have been investigated using powerful deep learning tools in the channel decoding \cite{Nachmani--Marciano--Lugosch--Gross--Burshtein--Beery}, \cite{Liang--Shen--Wu2018}, signal detection in MIMO systems  \cite{Yan--Long--Wang--Fu--Ou--Liu2017}, \cite{Samuel--Diskin--Wiesel2017}, and signal detection in chemical communications \cite{Farsad--Goldsmith2017}, \cite{Farsad--Goldsmith2018}.
A MIMO deep detection network  in the MIMO channel is noteworthy \cite{Samuel--Diskin--Wiesel2017}.
This detection network applies a deep unfolding approach that transforms a computationally intractable probabilistic model into a deep neural network by unfolding iterative calculations into neural-network layers (NNLs) \cite{Hershey--Roux--Weninger2014}.
Embedding the existing mathematical methods into black-box-like deep neural networks improves the accuracy and reduces complexity.
In this paper, we adopt the deep unfolding approach in the detection networks of the MIMO DF relay channel under three scenarios related to  the knowledge of the SR channel.
\subsection{Contributions}

The primary contributions are summarized as follows:
\begin{itemize}
\item With the instantaneous CSI of the SR link, a {\it detection network with SR channel (DNwSRC)} is proposed applying deep unfolding approach when the ML detector is applied at the relay.
The DNwSRC is trained using large numbers of randomly distributed channels.
This detection network detects signals on changing channels and requires much lower complexity compared to the suboptimal NML detector while exhibiting a fine performance. 

\item Applying an equivalent SR channel with the average error probability at the relay to the DNwSRC, a {\it detection network with relay error probability (DNwREP)} is proposed, which only requires the statistical CSI of the SR link.

\item Without any knowledge of the SR channel, a suboptimal exhaustive search detection algorithm called {\it NML with relay signal distance (NMLwRSD)} is proposed by considering the squared signal distance at the relay (Proposition \ref{prop:nmlwrsd}). 
To the best of our knowledge, the optimal or suboptimal detector  in this case has not been proposed in the literature.
This algorithm exhibits much better performance compared to the existing MD detector.  
Moreover, based on the NMLwRSD detector, a {\it detection network with relay signal distance (DNwRSD)} and a simplified DNwRSD (sDNwRSD) are proposed.
The DNwRSD and sDNwRSD achieve excellent performance without any knowledge of the SR channel.

\item 
To evaluate the performance of the NML-based detection networks (NMLDNs),  detection algorithms of {\it semidefinite relaxation (SDR) with SR channel (SDRwSRC)}, {\it SDR with relay error probability (SDRwREP)}, and {\it SDR with relay signal distance (SDRwRSD)}  are introduced as the SDR versions of the NML, NMLw2PEP, and NMLwRSD  detectors, respectively. 
The NML-based SDR (NMLSDR) detectors exhibit relatively good performance with polynomial complexity, and are thus suitable choices for the MIMO DF relay channel without enough training data.

\item Additionally, new linear detectors based on the zero gradient (ZG) of the metrics in the NML, NMLw2PEP, and NMLwRSD detectors are proposed (Propositions \ref{prop:ldwsrc}, \ref{prop:ldwrep}, and \ref{prop:ldwrsd}).
The NML-based ZG (NMLZG) detectors achieve much better performance than the existing linear ZFwMRC detector.

\item 
For various detection algorithms at the relay (DetR), we present the corresponding equivalent SR channel matrix such that the above-mentioned   detection algorithms at the destination (DetD) can be implemented  for any DetR.

\item We present and compare some DetR-DetD methods based on the characteristics in error probability and detection complexity in the MIMO relay channel.
This provides the directions for designing the system configuration.
\end{itemize}

The remainder of the paper is organized as follows. 
In the next section, we formally introduce the MIMO DF relay channel and its equivalent real model.
The main part of this paper is presented in Section III.
The NML-based detection networks (NMLDNs) such as the DNwSRC, DNwREP, DNwRSD, and sDNwRSD are proposed with various conditions of the knowledge of the SR channel.
In Section IV, the training and detection details for the proposed NMLDNs  are introduced.
For comparison, the SDR detectors based on the NML detection algorithms are proposed in Section V.
Various DetR-DetD methods are presented based on the required error probability and complexity after introducing the DetR and their corresponding equivalent SR channels in Section VI.
The main results are evaluated using TensorFlow and Matlab, and are detailed in Section VII.
Finally, the conclusions are given in Section VIII.

\subsection{Notations}
Throughout the paper, we use the following notations.
The superscript $(\cdot)^{T}$ denotes the transpose of a matrix;
$\tr(\cdot)$ denotes the trace of a matrix; 
$\Re(\cdot)$ and $\Im(\cdot)$ denote the real and imaginary parts of a complex number, respectively; 
$I_n$ denotes the $n \times n$ identity matrix 
(where the subscript $n$ is
omitted when it is irrelevant or clear from the context);
$\Complex^{n\times m}$ denotes a set of $n\times m$ complex
matrices; 
for $A\in\mathbb{C}^{n\times m}$,
$A\sim\CN(0,\sigma^2I_{nm})$ denotes that the elements of
$A$ are i.i.d. circularly
symmetric complex Gaussian random variables with zero mean and variance $\sigma^2$, and $B\sim\N(0,\sigma^2I_{nm})$ denotes that $B\in\mathbb{R}^{n\times m}$ is a real Gaussian random matrix with zero mean and covariance matrix $\sigma^2I_{nm}$;
$\diag(\cdot)$ denotes a block diagonal matrix with the entries on its main diagonal;
$[\cdot]_{i:j,k:l}$ means a matrix consisting of the entries from the $i$th row to the $j$th row, and from the $k$th column to the $l$th column in the original matrix. 
\section{System Model}\label{sec:sys_model}
A half-duplex DF relay channel with one source, one destination, and
one relay is considered.
It is assumed that the relay knows the CSI of
the SR channel, and the destination knows the
CSIs of  the SD and RD links.
In the first phase, the source with $N$ transmit antennas broadcasts $N$  independently and uniformly distributed complex signals $\xv^{C}=[x_1^{C},\dots,x_N^{C}]^T$ to the relay and the destination, where $\Re\{x_{i}^C\}\in \mathcal{A}, \Im\{x_{i}^C\}\in \mathcal{A}, i=1,\dots,N$, and $\mathcal{A}\in\{+1,-1\}$.
Subsequently, the received signals at the relay with $N_{\rt}$ receiving antennas can be written as 
\begin{align}\label{eq:srC}
	\yv_{\sr}^{C} = H_{\sr}^{C}\xv^{C}+\zv_{\sr}^{C}
\end{align}
where $H_{\sr}^{C}\in\Complex^{N_{\rt}\times  N}$ is the channel coefficient matrix of the SR link and  $\zv_{\sr}^C\sim\CN(0,\sigma^2I_{N_{\rt}})$ is the noise term at the relay. 
Simultaneously, the destination receives the signal transmitted from the source as
\begin{align}\label{eq:sdC}
	\yv_{\sd}^{C} &= H_{\sd}^{C}\xv^{C}+\zv_{\sd}^{C}
\end{align}
where  $H_{\sd}^{C}\in\Complex^{N_{\dt}\times  N}$ is the channel coefficient matrix of the SD link and  $\zv_{\sd}^C\sim\CN(0,\sigma^2I_{N_{\dt}})$ is the noise term at the destination in the first phase. 
In the second phase, the relay decodes the received signal and forwards the decoded signal $\xv_{\rt}^{C}=[x_{1{\rt}}^C,\dots,x_{N{\rt}}^C]^T$, $\Re\{x_{i{\rt}}^C\}\in \mathcal{A}, \Im\{x_{i\rt }^C\}\in \mathcal{A}, i=1,\dots,N$ to the destination. The received signal at the destination with $N_{\dt}$ receiving antennas in the second phase is 
\begin{align}\label{eq:rdC}
	\yv_{\rd}^{C} &= H_{\rd}^{C}\xv_{\rt}^{C}+\zv_{\rd}^{C}
\end{align}
where $H_{\rd}^{C}\in\Complex^{N_{\dt}\times  N}$  is the channel coefficient matrix of the RD link and  $\zv_{\rd}^C\sim\CN(0,\sigma^2I_{N_{\dt}})$ is the noise term at the destination in the second phase. 

\begin{figure}[t]
\centering{\includegraphics[width=3.5in]{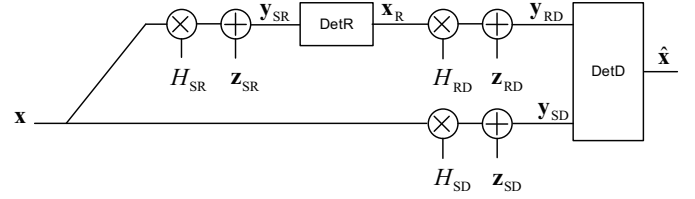}}
\caption{The equivalent real MIMO DF relay system model. 
}\label{fig:DF}
\end{figure}

To simplify the expressions, we convert the complex system model to a real system model. 
Let $\xv=\begin{bmatrix}\Re\{\xv^{C}\} \\\Im\{\xv^{C}\}\end{bmatrix}$,
 $\xv_{\rt}=\begin{bmatrix}\Re\{\xv^{C}_{\rt}\} \\\Im\{\xv^{C}_{\rt}\}\end{bmatrix}$, $\yv_{kl}=\begin{bmatrix}\Re\{\yv_{kl}^{C}\} \\\Im\{\yv_{kl}^{C}\}\end{bmatrix}$, 
$\zv_{kl}=\begin{bmatrix}\Re\{\zv_{kl}^{C}\} \\\Im\{\zv_{kl}^{C}\}\end{bmatrix}$, 
and $H_{kl} = \begin{bmatrix}\Re\{H_{kl}^{C}\}&-\Im\{H_{kl}^{C}\} \\ \Im\{H_{kl}^{C}\}&\Re\{H_{kl}^{C}\}\end{bmatrix}$ 
for $kl=\{\sr,\sd,\rd\}$.
Then an equivalent real system model is written as 
\begin{align}
	\yv_{\sr} &= H_{\sr}\xv~~+\zv_{\sr}\label{eq:sr}\\
	\yv_{\sd} &= H_{\sd}\xv~~+\zv_{\sd}\label{eq:sd}\\
	\yv_{\rd} &= H_{\rd}\xv_{\rt}+\zv_{\rd} \label{eq:rd}
\end{align}
where $\xv,\xv_{\rt}\in\mathcal{A}^{2N}$, $\zv_{\sr}\sim\mathcal{N}(0,\frac{1}{2}\sigma^2I_{2N_{\rt}})$, and $\zv_{kl}\sim\mathcal{N}(0,\frac{1}{2}\sigma^2I_{2N_{\dt}})$ for $kl=\{\sd,\rd\}$.
Further, the signal-to-noise ratios (SNRs) at the relay and the destination are linearly proportional to $\rho=\frac{2N}{\sigma^2}$.
The equivalent real system model is depicted in Fig. \ref{fig:DF}, where ``DetR'' and ``DetD'' represent the detection algorithms at the relay and the destination, respectively.
Various types of DetD and DetR will be introduced in the following sections.

\section{Deep Learning Detection Networks at Destination}\label{sec:DNaD}

Consider the MIMO relay channel in Fig. \ref{fig:DF}. 
Due to the probabilistic distribution of the noise terms $\zv_{\sr}$, $\zv_{\sd}$, and $\zv_{\rd}$, the optimal detection method is the ML detector that finds a $\hat{\xv}\in \mathcal{A}^{2N}$ that maximizes $p(\yv_{\sd},\yv_{\rd}|\xv, H_{\sd}, H_{\rd}, H_{\sr})$ for the uniformly distributed $\xv$.

As detection errors may exist in the relay, the detection method for the DF relay channel is different from that of the MIMO channel.
Error probabilities in the relay should be considered when the CSIs for the SR, SD, and RD links are known.
This means that both  signals possibly transmitted from the source and from the relay must be considered in the DetD.
This results in a much higher detection complexity in the  MIMO DF relay channel, approximately the square of the computational complexity of the  point-to-point MIMO channel detection.
The exhaustive search algorithms such as the ML and NML detectors cannot be applied in the MIMO DF relay channel  with large numbers of antennas.
Moreover, without the CSI of the SR link,  there is no existing optimal or suboptimal detection algorithm that we can refer to.
To address these problems, this section explores deep learning detection algorithms for three cases of the knowledge of the SR channel in the destination: 1) with the instantaneous CSI of the SR link; 2) with the statistical CSI of the SR link; 3) without the CSI of the SR link.

\subsection{With Instantaneous CSI of SR Link}\label{sec:wcsi}
With the full CSIs of the SR, SD, and RD links, the ML detection that maximizes the probability $p(\yv_{\sd},\yv_{\rd}|\xv,H_{\sd},H_{\rd}, H_{\sr})$ for the real system model in (\ref{eq:sr})-(\ref{eq:rd}) can be written as
\begin{align}\label{eq:MIMOML}
\hat{\xv}
&=\arg\max_{\xv\in \mathcal{A}^{2N}}p(\yv_{\sd},\yv_{\rd}|\xv,H_{\sd},H_{\rd}, H_{\sr})\notag\\
&=\arg\max_{\xv \in \mathcal{A}^{2N}}p\,(\yv_{\sd}|\xv,H_{\sd})\notag\\[-0.5em]
&\qquad\qquad\quad\cdot\sum_{\xv_{\rt}\in \mathcal{A}^{2N}}p\,(\yv_{\rd}|\xv_{\rt},H_{\rd} )   P_{\sr}(\xv_{\rt}|\xv, H_{\sr})\notag\\
&=\arg\max_{\xv \in \mathcal{A}^{2N}}\!\! \Bigg\{\!-\big\|\yv_{\sd}-H_{\sd} \xv\big\|^2\notag\\[-0.5em]
&\qquad\qquad\qquad+\!\sigma^2\!\ln \!\!\!\sum_{\xv_{\rt}\in \mathcal{A}^{2N}} \!\!\!\!\exp\!\bigg(\!-\frac{\big\|\yv_{\rd}-H_{\rd} \xv_{\rt}\big\|^2}{\sigma^2}\notag\\[-1em]
&\qquad\qquad\qquad\qquad\qquad\qquad\quad~+\!\ln P_{\sr}(\xv_{\rt}|\xv, H_{\sr})\Big]\bigg)\Bigg\}
\end{align}
where $P_{\sr}(\xv_{\rt}|\xv, H_{\sr})$ is the probability that the relay detects the received signal to $\xv_{\rt}$ when the source transmits $\xv$.
Since it is highly difficult to derive $P_{\sr}(\xv_{\rt}|\xv, H_{\sr})$ in MIMO systems \cite{MIMO_boundary}, the pair-wise error probability (PEP) between $\xv$ and $\xv_{\rt}$, $P_{\sr}(\xv \to \xv_{\rt}|H_{\sr})$ is used. 
Moreover, applying the widely-used max-log approximation $\ln\sum_{i}\exp(x_i)\approx \max_{i}x_i$  \cite{Hochwald--Brink2003}-\cite{Ju--Kim2009}, the near-ML (NML) detector was proposed in \cite{Jin2011ieice} as
\begin{align}\label{eq:MIMOnearML}
\hat{\xv}
&=\arg\min_{\xv\in \mathcal{A}^{2N}}  \min_{\xv_{\rt} \in \mathcal{A}^{2N}}
\Big\{\big\|\yv_{\sd}\!-\!H_{\sd} \xv\big\|^2 
\!+\!\big\|\yv_{\rd}\!-\!H_{\rd}\xv_{\rt}\big\|^2\notag\\
&\qquad\qquad\qquad\qquad\quad~- \sigma^2\ln P_{\sr}(\xv\!\rightarrow\! \xv_{\rt}|H_{\sr})\Big]\Big\}
\end{align}
where $P_{\sr}(\xv\to\xv_{\rt}|H_{\sr})=1 $ for $\xv_{\rt}=\xv$;  otherwise, $P_{\sr}(\xv\to\xv_{\rt}|H_{\sr})$ is the PEP between $\xv$ and $\xv_{\rt}$ for the ML detector at the relay (MLaR) \cite{Jin2011ieice} written as
\begin{align}\label{eq:pepR}
P_{\sr}(\xv\to\xv_{\rt}|H_{\sr})
=Q\bigg(\sqrt{\frac{1}{2\sigma^2}\|{H}_{\sr}(\xv-\xv_{\rt})\|^2}\bigg)
\end{align}
and $Q(x)=\frac{1}{\sqrt{2\pi}}\int_x^{\infty}e^{-\frac{y^2}{2}}dy$.

To detect the transmitted signal,  the NML detector requires  $|\mathcal{A}|^{4N}$ times of the calculation for the metric in \eqref{eq:MIMOnearML}.
This exhaustive search detection algorithm cannot be used in practice.
By unfolding the iterations of the projected gradient descent method to the layered neural networks as the deep MIMO detection in \cite{Samuel--Diskin--Wiesel2017},  a deep learning algorithm is proposed to approximate the NML detector.
This is described in detail in the following steps.
\begin{itemize}
\item
 Projected gradient descent method 

The projected gradient descent method is based on the gradient of the metric in the original  exhaustive search detection algorithm.
However, the metric in \eqref{eq:MIMOnearML} itself is unsuitable for the gradient descent method due to the complicated gradient of the function $\ln Q(x)$. 
Instead,  we use an approximation given in \cite{Jin2011ieice} 
\begin{align}\label{eq:Qapprox}
	&-\lim_{\sigma^2\to 0} \sigma^2 \ln Q\bigg(\sqrt{\frac{1}{2\sigma^2}\|{H}_{\sr}(\xv-\xv_{\rt})\|^2}\bigg)\notag\\
	&=\frac{1}{4}\|{H}_{\sr}(\xv-\xv_{\rt})\|^2.
\end{align}
Then the detection metric in \eqref{eq:MIMOnearML} becomes a quadratic function of $\underline{\xv}$:
\begin{align} 
m(\underline{\xv})
&=\big\|\yv_{\sd}-H_{\sd} \xv\big\|^2 +\big\|\yv_{\rd}-H_{\rd}\xv_{\rt}\big\|^2 \notag\\
&\qquad+\frac{1}{4}\big\|{H}_{\sr}(\xv-\xv_{\rt})\big\|^2\label{eq:metric_NML}\\[0.5em]
&=\|\underline{\yv}-H_{\dt}\underline{\xv}\|^2+\|H_{\rt}\underline{\xv}\|^2\label{eq:metric}
\end{align}
where $\underline{\xv}
=\begin{bmatrix}\xv~\\\xv_{\rt}\end{bmatrix}$, 
$\underline{\yv}
=\begin{bmatrix}\yv_{\sd}\\\yv_{\rd}\end{bmatrix}$, $H_{\dt}=\diag\big(H_{\sd},\,H_{\rd}\big)$,  and $H_{\rt}=\frac{1}{2}\begin{bmatrix}{H}_{\sr} & -{H}_{\sr}\end{bmatrix}$.
An optimization problem is established as 
\begin{align}\label{eq:opt}
	&\text{ minimize } m(\underline{\xv})\notag\\
	&\text{ subject to } \underline{\xv} \in \mathcal{A}^{4N}.
\end{align}

Applying the projected gradient descent method to the nonconvex optimization problem in \eqref{eq:opt}, an update in the $k$th iteration is written as 
\begin{align}\label{eq:PGD_relay}
	\hat{\underline{\xv}}_{k}
	&\!=\!\phi\Big(\hat{\underline{\xv}}_{k-1}-\delta_k\nabla m(\underline{\xv})\Big|_{\underline{\xv}=\hat{\underline{\xv}}_{k-1}}\Big)\notag\\
	&\!=\!\phi\Big(\hat{\underline{\xv}}_{k\!-\!1}\!+\!\delta_k 2H_{\dt}^T\underline{\yv}\!- \!\delta_k 2H_{\dt}^TH_{\dt}\hat{\underline{\xv}}_{k\!-\!1}\!-\!\delta_k 2H_{\rt}^TH_{\rt}\hat{\underline{\xv}}_{k\!-\!1}\Big) 
\end{align}
where $\phi(\cdot)$ is a nonlinear projection operator, e.g.,  
$\phi(x)=\sign(x)$ for $x\in\mathcal{A}=\{+1,-1\}$,
 $\hat{\xv}_{k-1}$ is the estimate in the $(k-1)$th iteration,  $\delta_{k}$ is a step size in the $k$th iteration for $k=1,\dots, L$, 
  $\nabla m(\underline{\xv})=-2H_{\dt}^T\underline{\yv}+ 2H_{\dt}^TH_{\dt}\underline{\xv}+2H_{\rt}^TH_{\rt}\underline{\xv}$,~ $H_{\dt}^T\underline{\yv}=\begin{bmatrix}  
	H_{\sd}^T\yv_{\sd}\\
	H_{\rd}^T\yv_{\rd} \end{bmatrix}$,\\
	$H_{\dt}^TH_{\dt}\hat{\underline{\xv}}_{k-1} 
	=\begin{bmatrix}  
	H_{\sd}^TH_{\sd}\hat{\xv}_{k-1}~~~~\\
	H_{\rd}^TH_{\rd}\hat{\xv}_{\rt,k-1}\end{bmatrix}$, and  
	$H_{\rt}^TH_{\rt}\hat{\underline{\xv}}_{k-1} 
	=\frac{1}{4}\begin{bmatrix}  
	~\,H_{\sr}^TH_{\sr}(\hat{\xv}_{k-1}-\hat{\xv}_{\rt,k-1})\\
	-H_{\sr}^TH_{\sr}(\hat{\xv}_{k-1}-\hat{\xv}_{\rt,k-1})\end{bmatrix}$.
\begin{figure}[t]
\centering{\includegraphics[width=3.3in]{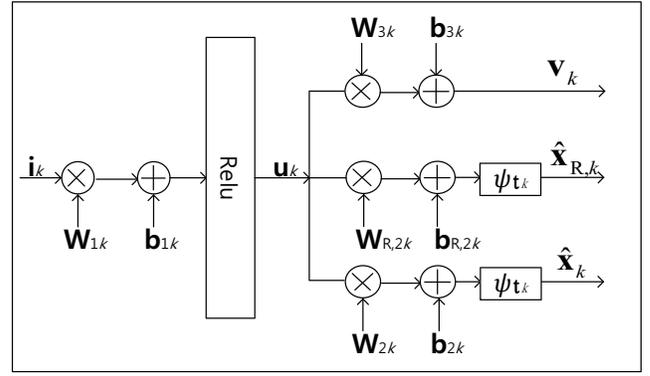}}
\caption{A single NNL in the detection network. 
}\label{fig:DFR_est_net}
\end{figure}
\begin{figure*}[h]
\centering{\includegraphics[width=6.7in]{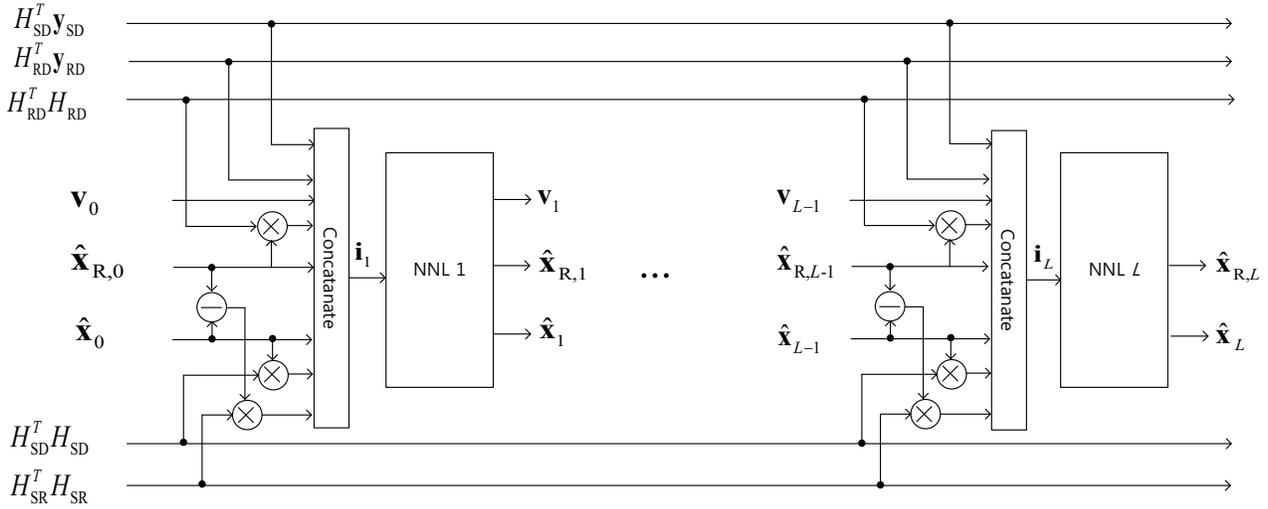}}
\caption{The DNwSRC in the MIMO DF relay channel. 
}\label{fig:det_net}
\end{figure*}
$~$\\

\item Unfolding iterations 

The $L$ iterations are unfolded to the $L$ neural-network layers (NNLs).
The update in the $k$th iteration  in \eqref{eq:PGD_relay} is reflected in the input of the $k$th NNL, i.e.,
\begin{align}\label{eq:mimodf_input}
	\iv_k &=\begin{bmatrix}  
	H_{\sd}^T\yv_{\sd}\\
	H_{\rd}^T\yv_{\rd}\\
	\vv_{k-1}\\
	\hat{\xv}_{k-1}\\
	\hat{\xv}_{\rt,k-1}\\
	H_{\sd}^TH_{\sd}\hat{\xv}_{k-1}~~~~\\
	H_{\rd}^TH_{\rd}\hat{\xv}_{\rt,k-1}~~~~\\
   {H}_{\sr}^T{H}_{\sr}(\hat{\xv}_{k-1}-\hat{\xv}_{\rt,k-1})
	 \end{bmatrix}.
\end{align}
Due to the structure of the NNL in Fig. \ref{fig:DFR_est_net}, the repeated part and the constant values in \eqref{eq:PGD_relay} are ignored in \eqref{eq:mimodf_input}.
Unlike the deep MIMO detection \cite{Samuel--Diskin--Wiesel2017}, the output of each layer includes an auxiliary vector $\vv$ and the desired signal $\xv$ as well as a help signal $\xv_{\rt}$.
In detail, the main parts in the $k$th NNL include
\begin{align*}
	  \uv_k&=\relu(\Wv_{1k}\iv_k~~+\bv_{1k})\\
\hat{\xv}_k&=\psi_{\tv_k}~(\Wv_{2k}\uv_k~~+\bv_{2k})\\
\hat{\xv}_{\rt,k}&=\psi_{\tv_k}~(\Wv_{\rt,2k}\uv_k+\bv_{\rt,2k})\\
     \vv_k&=\qquad\,\Wv_{3k}\uv_k~~+\bv_{3k}
\end{align*}
where $\relu(x)=\max(0,x)$ and $\psi_{\tv_k}$ is a element-wise soft decision operator. 
We adopt $\psi_{t_k}(x)=-1+\frac{\relu(x+t_k)}{|t_k|}-\frac{\relu(x-t_k)}{|t_k|} $ in \cite{Samuel--Diskin--Wiesel2017} for $x\in \{1,-1\}$. 
As shown in Fig. \ref{fig:det_net}, the entire detection network includes $L$ NNLs, where the outputs of the previous layer, $\vv,\xv_{\rt}$, and $\xv$,  combined with the observation and side information $H_{\sd}^T\yv_{\sd},H_{\rd}^T\yv_{\rd}$, $H_{\sd}^TH_{\sd}$, $H_{\rd}^TH_{\rd}$, and ${H}_{\sr}^T{H}_{\sr}$, enter the next layer. 
In the last layer, the final decision is made as $\hat{\xv}=\phi(\hat{\xv}_L)$. 
To improve performance, we adopt the residual learning \cite{He--Zhang--Ren--Sun2016}, i.e., applying a weighted average of the previous output and the current output to the current output \cite{Samuel--Diskin--Wiesel2017}.

$~$
\item Learning algorithm

The learning algorithm of the detection network at the destination in Fig. \ref{fig:det_net} is denoted as
\begin{align}\label{eq:learning_algo_nml}
\hat{\xv}= g_{\theta} (\yv_{\sd},\yv_{\rd}, H_{\sd}, H_{\rd}, H_{\sr}) 
\end{align}
where 
\begin{align}\label{eq:param}
{\bf \theta}=\{&\Wv_{1k}, \bv_{1k},\Wv_{2k},\bv_{2k},\Wv_{\rt,2k},\bv_{\rt,2k},\Wv_{3k},\bv_{3k}, \tv_{k},\notag\\
& \qquad k=1,\dots, L\}
\end{align}
is a set of parameters that is trained during the training phase.

$~$
\item Loss function 

To train the learning algorithm, $g_{\theta}$, in \eqref{eq:learning_algo_nml}, we can use the reference signal $\xv$ and  the observation $(\yv_{\sd},\yv_{\rd})$ with the side information ($H_{\sd}, H_{\rd}, H_{\sr}$) as the training data.
From the projected gradient descent method in \eqref{eq:PGD_relay} and the NNL in Fig. \ref{fig:DFR_est_net}, we can find that $\xv$ and $\xv_{\rt}$ affect each other in each iteration or layer.
Thus, setting $\xv_{\rt}$ as another reference signal helps to improve the accuracy.
Since the training phase is a preprocessing step, $\xv_{\rt}$ can be known as a reference signal before training.

Combining all outputs of the NNLs \cite{Samuel--Diskin--Wiesel2017}, two possible loss functions can be used. 

1) When both $\xv$ and $\xv_{\rt}$ are known as the reference signal, the loss function $l_1$ can be employed as
\begin{align}\label{eq:l1}
	l_1(\underline{\xv};\hat{\underline\xv}_{\theta})=\sum_{k=1}^L \log(k+1)\frac{\|\underline{\xv}-\hat{\underline\xv}_k\|^2}{\|\underline{\xv}-\tilde{\underline\xv}\|^2} 
\end{align}
where $\underline{\xv}=\begin{bmatrix}\xv~\\ \xv_{\rt} \end{bmatrix} $, $\hat{\underline{\xv}}_k=\begin{bmatrix}\hat{\xv}_k~\\ \hat{\xv}_{\rt,k} \end{bmatrix} $, and $\tilde{\underline\xv}= (H_{\dt}^TH_{\dt}+H_{\rt}^TH_{\rt})^{-1}H_{\dt}^T \underline{\yv}$, which will be derived in Proposition \ref{prop:ldwsrc}.

2) When only $\xv$ is known as the reference signal, the transmitted signal from the relay, $\xv_{\rt}$, cannot be used for training, and the loss function $l_2$ is used instead:
\begin{align}\label{eq:l2}
	l_2(\xv;\hat{\xv}_{\theta})=\sum_{k=1}^L \log(k+1)\frac{\|\xv-\hat{\xv}_k\|^2}{\|\xv-\tilde{\xv}\|^2}
\end{align}
where $\tilde{\xv}=[\,\tilde{\underline\xv}\,]_{1:2N}$.
\end{itemize}
This detection method is called  a {\it detection network  with SR channel (DNwSRC)} in the MIMO DF relay channel.
\begin{proposition}\label{prop:ldwsrc}
Obtaining the zero gradient point for the convex function in \eqref{eq:metric}, i.e.,
$$\nabla m(\underline{\xv})=-2H_{\dt}^T\underline{\yv}+ 2H_{\dt}^TH_{\dt}\underline{\xv}+2H_{\rt}^TH_{\rt}\underline{\xv}=0,$$
a linear receiver in the MIMO DF relay channel can be obtained as  
\begin{align}\label{eq:lrwsrc}
\tilde{\underline{\xv}}=(H_{\dt}^TH_{\dt}+H_{\rt}^TH_{\rt})^{-1}H_{\dt}^T \underline{\yv} 
\end{align}
where 
$H_{\dt}^TH_{\dt}=\diag\big(H_{\sd}^TH_{\sd}, H_{\rd}^TH_{\rd}\big)$,
$H_{\rt}^TH_{\rt}=\frac{1}{4}\begin{bmatrix}  
	H_{\sr}^TH_{\sr} & -H_{\sr}^TH_{\sr}\\
	-H_{\sr}^TH_{\sr}& H_{\sr}^TH_{\sr} \end{bmatrix}$,
and 
$H_{\dt}^T\underline{\yv}=\begin{bmatrix}  
	H_{\sd}^T\yv_{\sd}\\
	H_{\rd}^T\yv_{\rd} \end{bmatrix}$.
Subsequently, a new linear detector for the desired signal $\xv$ is derived as 
\begin{align}\label{eq:ldwsrc}
	\hat{\xv}
	&=\phi\big(\tilde{\xv}\big)
\end{align}
where $\tilde{\xv}=[\,\tilde{\underline\xv}\,]_{1:2N}$.
This is called a detector of  zero gradient with SR channel (ZGwSRC).
\end{proposition}

\subsection{With Statistical CSI of SR Link}\label{sec:wscsi}
In this section, we handle the case where only the statistical CSI of the SR link is known at the destination for the Rayleigh fading  SR channel   $H_{\sr}^{C}\sim\CN(0,\sigma^2_{\sr}I_{N_{\rt}N})$.
Since the instantaneous CSI of the SR link, $H_{\sr}$, is unknown at the destination, the exact PEP, $P_{\sr}(\xv\to\xv_{\rt}|H_{\sr})$, could not be applied  in \eqref{eq:MIMOnearML}.
Instead, the average PEP, $\bar{P}_{\sr}$, can be used. 
Subsequently, a NML detector with two-level-PEP (NMLw2PEP) \cite{Jin2014jcn} is written as  
\begin{align}\label{eq:MIMOnearML2PEP}
\hat{\xv}
&=\arg\min_{\xv\in \mathcal{A}^{2N}}  \min_{\xv_{\rt} \in \mathcal{A}^{2N}}\!
\Big\{\big\|\yv_{\sd}\!-\!H_{\sd} \xv\big\|^2 \!+\!\big\|\yv_{\rd}\!-\!H_{\rd}\xv_{\rt}\big\|^2\notag\\
&\qquad\qquad\qquad\qquad\qquad-\sigma^2\!\ln \bar{P}_{\sr}\Big\} \notag\\
&=\arg\min_{\xv\in \mathcal{A}^{2N}}  \min_{\xv_{\rt} \in \mathcal{A}^{2N}}\! \Big\{\big\|\yv_{\sd}\!-\!H_{\sd} \xv\big\|^2\! +\!\big\|\yv_{\rd}\!-\!H_{\rd}\xv_{\rt}\big\|^2 \notag\\
&\qquad\qquad\qquad\qquad\qquad+\sigma^2\!\ln P_e^{-1} \cdot {\bf 1}_{\xv_{\rt}\ne\xv}\Big\}
\end{align}
where $\bar{P}_{\sr}
=\begin{cases}1 & \text{~for~} \xv_{\rt}=\xv \\
P_e & \text{~for~} \xv_{\rt}\ne\xv,
\end{cases}$~~
${\bf 1}_{\xv_{\rt}\ne\xv}
=\begin{cases}
0 & \text{~for~} \xv_{\rt}=\xv~\\
1 & \text{~for~} \xv_{\rt}\ne\xv,
\end{cases}$~~ 
$P_e\in(0,1]$ can be expressed as 
$P_e=\gamma_{\sr}^{-d_{\rt}}$, $\gamma_{\sr}=\frac{2N\sigma_{\sr}^2}{\sigma^2}$ is the average SNR,  and $d_{\rt}=N_{\rt}$ is the diversity order when the ML is used at the relay \cite[Lemma 1]{Jin2014jcn}.

To apply the projected gradient descent method, we need to take a gradient for the metric in \eqref{eq:MIMOnearML2PEP}, but ${\bf 1}_{\xv_{\rt}\ne\xv}$ is inappropriate for taking the gradient. 
Considering the nearest points $\xv'$ and $\xv'_{\rt}$ satisfying $\|\xv'-\xv'_{\rt}\|^2=4=\displaystyle\min_{\xv_{\rt}\ne \xv}\|\xv-\xv_{\rt}\|^2$, we apply $\frac{1}{4}\|\xv-\xv_{\rt}\|^2$ instead of ${\bf 1}_{\xv_{\rt}\ne\xv}$. 
Then the metric for the NMLw2PEP can be rewritten as
\begin{align}\label{eq:metric_NML2PEP}
m(\underline{\xv})
&=\big\|\yv_{\sd}-H_{\sd} \xv\big\|^2 +\big\|\yv_{\rd}-H_{\rd}\xv_{\rt}\big\|^2 \notag\\
&\qquad+\frac{1}{4} \sigma^2\!\ln P_e^{-1} \cdot\|\xv-\xv_{\rt}\|^2
\end{align}
where $\underline{\xv}
=\begin{bmatrix}\xv^T&\xv_{\rt}^T\end{bmatrix}^T$. 

Substituting ${H}_{\sr}^T{H}_{\sr}=  \sigma^2\!\ln P_e^{-1} \cdot I$ 
into \eqref{eq:mimodf_input}
and training the detection network in Fig. \ref{fig:det_net} to minimize the loss function in \eqref{eq:l1} or \eqref{eq:l2},  a new deep learning detection algorithm namely {\it detection network with   relay error probability (DNwREP)} is achieved.
The loss functions are normalized using a linear receiver in the following proposition.
\begin{proposition}\label{prop:ldwrep}
A linear receiver $\tilde{\underline{\xv}}$ is obtained by substituting 
${H}_{\sr}^T{H}_{\sr}= \sigma^2\!\ln P_e^{-1} \cdot I$ into \eqref{eq:lrwsrc}. 
Making decisions to the desired part, i.e., $\hat{\xv}=\phi\big([\,\tilde{\underline\xv}\,]_{1:2N}\big)$, a detector of zero gradient with relay error probability (ZGwREP) is derived.
\end{proposition}
\subsection{Without CSI of SR Link}\label{sec:wocsi}

Without the CSI of the SR link, the minimum distance (MD) detector \cite{Jin2011ieice} can first be considered.
The MD detector ignores the detection errors at the relay although the error may occur.
Setting $\xv_{\rt}=\xv$ in \eqref{eq:MIMOnearML}, the MD detection algorithm is written as
\begin{align}\label{eq:MD}
\hat{\xv}
&=\arg\min_{\xv \in \mathcal{A}^{2N}}  
\big\{\big\|\yv_{\sd}-H_{\sd} \xv\big\|^2
+ \big\|\yv_{\rd}-H_{\rd} \xv\big\|^2\big\} 
\end{align}
and its linear detection version is 
\begin{align}\label{eq:ldwmrc}
	\hat{\xv}
	=\phi\big(\big(H_{\sd}^T H_{\sd}+H_{\rd}^TH_{\rd}\big)^{-1} (H_{\sd}^T\yv_{\sd}+H_{\rd}^T\yv_{\rd})\big)
\end{align}
which was proposed in \cite{Chalise--Vandendorpe2009}. 
We call it a detector of ZF with the MRC (ZFwMRC).

Since the error probability at the relay  is not considered in the MD and the ZFwMRC detectors, they exhibit poor performance \cite{Jin2011ieice}.
From the metrics for the NML and NMLw2PEP detectors in \eqref{eq:metric_NML} and \eqref{eq:metric_NML2PEP}, we can find that 
the influence of the SR link in the metrics exists only when $\xv_{\rt}\ne\xv$.
Due to the similar reason presented in Section \ref{sec:wscsi}, $\frac{1}{4}\|\xv-\xv_{\rt}\|^2$ can  represent the influence of the SR link well regardless of the CSI of the SR link.
Meanwhile, adding $\frac{1}{4}\|\xv-\xv_{\rt}\|^2$ can be regarded as a regularization method in convex optimization problems \cite{Boyd--Vandenberghe2004}.
Subsequently, we propose a suboptimal detection algorithm in the following proposition.
\begin{proposition}\label{prop:nmlwrsd}
Without any knowledge of the SR link,  an NML with relay signal distance (NMLwRSD)  detector is proposed as
\begin{align}\label{eq:NMLwRSD}
\hat{\xv}
&=\arg\min_{\xv\in \mathcal{A}^{2N}}  \min_{\xv_{\rt} \in \mathcal{A}^{2N}}
\Big\{\big\|\yv_{\sd}\!-\!H_{\sd} \xv\big\|^2 \!+\!\big\|\yv_{\rd}\!-\!H_{\rd}\xv_{\rt}\big\|^2\notag\\
&\qquad\qquad\qquad\qquad \qquad+\frac{1}{4}\|\xv-\xv_{\rt}\|^2\Big\}.
\end{align}
\end{proposition}
This NMLwRSD detector will achieve improved performance by considering the influence of the SR channel, however, it requires exhaustive search for all possible signal sets.
Based on the NMLwRSD detector, we propose a new deep learning detection algorithm called {\it detection network with relay signal distance (DNwRSD)}, by substituting ${H}_{\sr}^T{H}_{\sr}=I$ into \eqref{eq:mimodf_input}.
This detection network is trained using the loss function  in \eqref{eq:l1} or \eqref{eq:l2}  normalized by a linear receiver proposed in Proposition \ref{prop:ldwrsd} at the end of this section.

Furthermore, observing  \eqref{eq:mimodf_input}, when ${H}_{\sr}^T{H}_{\sr}=I$, the last term becomes $\hat{\xv}_{k-1}-\hat{\xv}_{\rt,k-1}$ and can be represented by $\hat{\xv}_{k-1}$ and $\hat{\xv}_{\rt,k-1}$.
Thus, the input dimension can be reduced, i.e.,
$\iv_k$ becomes
\begin{align}\label{eq:mimodf_input_reduced}
	\iv_k &=\begin{bmatrix}  
	H_{\sd}^T\yv_{\sd}\\
	H_{\rd}^T\yv_{\rd}\\
	\vv_{k-1}\\
	\hat{\xv}_{k-1}\\
	\hat{\xv}_{\rt,k-1}\\
	H_{\sd}^TH_{\sd}\hat{\xv}_{k-1}~\\
	H_{\rd}^TH_{\rd}\hat{\xv}_{\rt,k-1}
	 \end{bmatrix}.
\end{align}
Substituting ${H}_{\sr}^T{H}_{\sr}=0$ into the detection network in Fig.\ref{fig:det_net},  a new deep learning detection algorithm, namely {\it simplified DNwRSD (sDNwRSD)} is obtained.
\begin{proposition}\label{prop:ldwrsd}
A linear receiver $\tilde{\underline{\xv}}$ is obtained by substituting 
${H}_{\sr}^T{H}_{\sr}=I$ into \eqref{eq:lrwsrc}. 
Making decisions for the desired part, i.e., $\hat{\xv}=\phi\big([\,\tilde{\underline\xv}\,]_{1:2N}\big)$,  a detector of zero gradient with relay signal distance (ZGwRSD) is derived.
\end{proposition}

%

\section{Training and Detection Details of NMLDNs}

\begin{algorithm}[t]
\caption{The training process of NMLDNs using the loss function $l_1$ in \eqref{eq:l1}}\label{alg:dnad_train}
\begin{algorithmic}[1]
\State Initialize $\hat{\xv}_0$, $\hat{\xv}_{\rt,0}$, $\vv_0$, the parameter set $\theta_0$ in \eqref{eq:param}, and $P_{\min}$

\For {iteration $n=1,2,\dots, N_{\text{iteration}}$ }

\State \parbox[t]{\dimexpr\linewidth-\algorithmicindent}{Generate $N_{\text{batch}}$ data samples of $\xv, \yv_{\sr},\yv_{\sd},\yv_{\rd}, H_{\sd}, H_{\rd}, H_{\sr}$ randomly according to their  distributions and obtain  $\xv_{\rt}$ by applying the DetR\strut}

\State \parbox[t]{\dimexpr\linewidth-\algorithmicindent}{Compute  $H_{\sd}^T\yv_{\sd}, H_{\rd}^T\yv_{\rd},  H_{\sd}^TH_{\sd}, H_{\rd}^T H_{\rd}$, and $ H_{\sr}^TH_{\sr}$\strut}

\For{layer $k=1,2,\dots, L$}
\State \parbox[t]{\dimexpr\linewidth-\algorithmicindent-\algorithmicindent}{Compute $\iv_k$ using $\hat{\xv}_{k-1}$, $\hat{\xv}_{\rt,k-1}$, $\vv_{k-1}$, $H_{\sd}^T\yv_{\sd}, H_{\rd}^T\yv_{\rd},  H_{\sd}^TH_{\sd}, H_{\rd}^T H_{\rd},  H_{\sr}^TH_{\sr}$\strut}

\State \parbox[t]{\dimexpr\linewidth-\algorithmicindent-\algorithmicindent}{Compute $\hat{\xv}_k$, $\hat{\xv}_{\rt,k}$, and $\vv_k$ using the parameters $\Wv_{1k}, \bv_{1k},\Wv_{2k},\bv_{2k},\Wv_{\rt,2k},$ $\bv_{\rt,2k},\Wv_{3k},\bv_{3k}, \tv_{k}$ in $\theta_{n-1}$\strut}
\EndFor

\State \parbox[t]{\dimexpr\linewidth-\algorithmicindent}{Update $\theta_{n-1}$ to $\theta_n$  with Adam optimizer \cite{Kingma--Ba2014} to minimize $l_{\text{ave}}=\frac{1}{N_{\text{batch}}}\sum_{m=1}^{N_{\text{batch}}}l_1(\underline{\xv}^m;\hat{\underline\xv}_{\theta}^m)$ where $l_1(\underline{\xv}^m;\hat{\underline\xv}_{\theta}^m)$ is the one  in \eqref{eq:l1} for $m$th batch\strut}
\State \parbox[t]{\dimexpr\linewidth-\algorithmicindent}{Determine $\hat{\xv}^m=[\hat{x}_1^m,\dots, \hat{x}_{2N}^m]^T=\phi(\hat{\xv}_L^m)$ for $m=1,\dots, N_{\text{batch}}$\strut}
\State Calculate $P_{b,n}=\frac{1}{2NN_{\text{batch}} }\sum_{m=1}^{N_{\text{batch}}}\sum_{i=1}^{2N} {\bf 1}_{x_i^m\ne \hat{x}_i^m} $

\If {$n\%N_{\text{eval}}==0$}
\State Calculate $P_b = \frac{1}{N_{\text{eval}}}\sum_{j=1}^{N_{\text{eval}}}P_{b,n-N_{\text{eval}}+j}$
\If {$P_b<P_{\min}$}
     \State $P_{\min }=P_b$
     \State Save parameters $\theta_n$ to $\theta^*$
\EndIf

\EndIf
\EndFor	
\State \textbf{return} $\theta^*$
\end{algorithmic}
\end{algorithm}
\begin{algorithm}
\caption{The detection stage of NMLDNs }\label{alg:dnad_alg}
\begin{algorithmic}[1]
\State Set $\hat{\xv}_0$, $\hat{\xv}_{\rt,0}$, and $\vv_0$ as the same as the ones in Algorithm \ref{alg:dnad_train}

\State Compute  $H_{\sd}^T\yv_{\sd}, H_{\rd}^T\yv_{\rd},  H_{\sd}^TH_{\sd}, H_{\rd}^T H_{\rd}$, and $ H_{\sr}^TH_{\sr}$

\For{layer $k=1,2,\dots, L$}
\State \parbox[t]{\dimexpr\linewidth-\algorithmicindent}{Compute $\iv_k$ using $\hat{\xv}_{k-1}$, $\hat{\xv}_{\rt,k-1}$, $\vv_{k-1}$, $H_{\sd}^T\yv_{\sd}, H_{\rd}^T\yv_{\rd},  H_{\sd}^TH_{\sd}, H_{\rd}^T H_{\rd},  H_{\sr}^TH_{\sr}$\strut}

\State \parbox[t]{\dimexpr\linewidth-\algorithmicindent}{Compute $\hat{\xv}_k$, $\hat{\xv}_{\rt,k}$, and $\vv_k$ using the final parameters $\theta^*$ trained in Algorithm \ref{alg:dnad_train}\strut}
\EndFor

\State \textbf{return} $\hat{\xv}=\phi(\hat{\xv}_L)$
\end{algorithmic}
\end{algorithm}

The training for the NMLDNs such as DNwSRC, DNwREP, DNwRSD, and sDNwRSD is implemented on the TensorFlow frameworks \cite{tensorflow} by applying the Adam optimizer, a variation of the stochastic gradient descent method \cite{Kingma--Ba2014}.
The equivalent real system in \eqref{eq:sr}-\eqref{eq:rd}  with independently and uniformly distributed input $\xv\in\mathcal{A}^{2N}$ is applied.
Using a $4N$ dimensional vector $\vv_k$, we have a $18N$ dimensional input vector $\iv_{k}$ in \eqref{eq:mimodf_input}.
The dimension of the input vector $\iv_{k}$ is reduced to $16N$ for the sDNwRSD.

To train the detection networks, $N_{\text{batch}}=5000$ data samples of $\xv, \yv_{\sr},\yv_{\sd},\yv_{\rd}, H_{\sr}, H_{\sd}$, and $H_{\rd}$ are randomly generated according to their  distributions in each iteration, and  $N_{\text{iteration}}=5\times 10^4$ iterations are implemented.
The training process of the NMLDNs with the loss function $l_1$ in \eqref{eq:l1} is shown in Algorithm \ref{alg:dnad_train}, where $H_{\sr}$ is the equivalent SR channel matrix depending on the type of NMLDN in Section \ref{sec:DNaD}, and $P_b = \frac{1}{N_{\text{eval}}}\sum_{j=1}^{N_{\text{eval}}}P_{b,n-N_{\text{eval}}+j}$ is the average bit error rate (BER) during $N_{\text{eval}}=50$ iterations.
A total of $2N\,N_{\text{eval}}N_{\text{batch}}$ bits can be used to evaluate the performance to well reflect the performance of BER$>\frac{100}{2N\,N_{\text{eval}}N_{\text{batch}}}=\frac{2}{N}\times 10^{-4}$.
Through the offline training process of Algorithm \ref{alg:dnad_train}, the final parameter set $\theta^*$ is determined.
Once  $\theta^*$ is determined, the transmitted signal $\xv$ can be detected in real time using Algorithm \ref{alg:dnad_alg}. 
The training  for the loss function $l_2$ in \eqref{eq:l2} can be performed similarly.

\section{SDR Approach in MIMO DF Relay Channels}\label{sec:sdr}
In this section, the SDR technique \cite{Jalden--Ottersten2008}, \cite{Luo--Ma--So--Ye--Zhang2010}  with polynomial complexity is applied in the MIMO DF relay channels as a performance comparison to the proposed NMLDNs.
We begin from the optimization problem in \eqref{eq:opt} that is a revised version of the NML detector in \eqref{eq:MIMOnearML}.
The optimization problem  for $\mathcal{A}=\{1,-1\}$ can equivalently be rewritten as 
\begin{align}\label{eq:opt2}
	&\text{ minimize } \tr(\Lv\Xv)\notag\\
	&\text{ subject to } [\Xv]_{ii}=1,~~ i=1,\dots,4N+1 \notag\\
	&\qquad\qquad~~  \Xv = \sv\sv^T
\end{align}
where
\begin{align}\label{eq:opt2_coef}
	&\Lv =
\begin{bmatrix}  
	H_{\dt}^TH_{\dt}+H_{\rt}^TH_{\rt} & -H_{\dt}^T\underline{\yv}\\
	-H_{\dt}^T\underline{\yv} & \underline{\yv}^T\,\underline{\yv}
\end{bmatrix}, 
\end{align}
$\sv\!=\!\begin{bmatrix}  
	\hat{\underline{\xv}}\\
	1
\end{bmatrix}$,
$\hat{\underline{\xv}}
\!=\!\begin{bmatrix}\hat{\xv}~\,\\\hat{\xv}_{\rt}\end{bmatrix}$, 
$\underline{\yv}
\!=\!\begin{bmatrix}\yv_{\sd}\\\yv_{\rd}\end{bmatrix}$, $H_{\dt}\!=\!\diag\big(H_{\sd},\,H_{\rd}\big)$, and  
$H_{\rt}^TH_{\rt}\!=\!\frac{1}{4}\!\begin{bmatrix}  
	~H_{\sr}^TH_{\sr} & -H_{\sr}^TH_{\sr}\\
	-H_{\sr}^TH_{\sr}& ~\,H_{\sr}^TH_{\sr} \end{bmatrix}$.
By replacing the last constraint $\Xv = \sv\sv^T$ with $ \Xv\succeq 0$ in the nonconvex optimization problem in \eqref{eq:opt2}, an SDR  problem is obtained as
\begin{align}\label{eq:opt_sdr}
	&\text{ minimize } \tr(\Lv\Xv)\notag\\
	&\text{ subject to } [\Xv]_{ii}=1, ~~i=1,\dots,4N \notag\\
	&\qquad\qquad~~  \Xv\succeq 0
\end{align}
which can be solved by standard
convex optimization techniques \cite{Boyd--Vandenberghe2004} or the CVX packages in MATLAB, e.g., \cite{Grant--Boyd2013}. 
Subsequently, the desired signal can be detected as 
\begin{align}
	\hat{\xv}=\sign([\Xv]_{1:2N,4N+1})
\end{align}
which is called a  detector of  SDR with SR channel  (SDRwSRC).
Since the relationship of $\Xv = \sv\sv^T=\begin{bmatrix}  
	\hat{\underline{\xv}}\,\hat{\underline{\xv}}^T & \hat{\underline{\xv}}\\
	\hat{\underline{\xv}}^T &1
\end{bmatrix}$ 
is not established in the problem of \eqref{eq:opt_sdr}, this SDR detector cannot achieve the same performance as the one in \eqref{eq:opt2}.
However, we can expect a fine performance similar to the case of the MIMO channel \cite{Jalden--Ottersten2008}, \cite{Luo--Ma--So--Ye--Zhang2010}, where the SDR detector achieves the same diversity order with the ML detector.

Substituting ${H}_{\sr}^T{H}_{\sr}=\sigma^2\ln P_e^{-1}\cdot I$ and ${H}_{\sr}^T{H}_{\sr}= I$ into \eqref{eq:opt2_coef}, and plugging \eqref{eq:opt2_coef} into \eqref{eq:opt_sdr}, two SDR detectors, an SDR with relay error probability (SDRwREP) and an SDR  with relay signal distance (SDRwRSD), are obtained.

\section{MIMO Relay System Configuration}\label{eq:relay_design}
In the previous sections, four types of detection algorithms at the destination  (DetD) are introduced when the ML detector is applied at the relay. 
They are the NML detectors such as NML, NMLw2PEP, and NMLwRSD;~ 
the NML-based ZG (NMLZG) detectors such as ZGwSRC, ZGwREP, and ZGwRSD;~
the NML-based detection networks (NMLDNs) such as DNwSRC, DNwREP, DNwRSD, and sDNwRSD;~
and the NML-based SDR detectors such as SDRwSRC, SDRwREP, and SDRwRSD.
Regardless of the DetD, various types of detectors can be used at the relay.
For different detection algorithms at the relay (DetR), the equivalent channel matrix of the SR link applied in the DetD in  Sections \ref{sec:wcsi}, \ref{sec:wscsi}, and \ref{sec:sdr} should be different. 
In the following subsections, we introduce some DetR and their equivalent SR channels, discuss the detection complexities of the DetR and DetD, and subsequently present various DetR-DetD methods according to the error performance and detection complexity.

\subsection{Detection Algorithms at the Relay (DetR)}\label{sec:DetR}

In this section, we briefly introduce some representative DetR and handle the corresponding equivalent SR channel $\tilde{H}_{\sr}$ applied in the DetD.
\begin{itemize}

\item[1)] The ML detector at the relay (MLaR)

The optimal MLaR is written as
\begin{align}\label{eq:mimoML}
\hat{\xv}
&=\arg\min_{\xv \in \mathcal{A}^{2N}} \big\|\yv_{\sr}-H_{\sr} \xv\big\|^2,
\end{align}
and the PEP between $\xv$ and $\xv_{\rt}$ is in \eqref{eq:pepR}.
Thus, the original ${H}_{\sr}$ is used in the detections at the destination.

\item[2)] The ZF detector at the relay (ZFaR)\footnote{The MMSE detector is also widely used in the MIMO channel. When $\sigma^2\to 0$, the MMSE detector is the same as the ZF detector.} 

The ZFaR \cite{Lupas--Verdu1989} is written as
\begin{align}
\hat{\xv}=\phi\big(\tilde{\xv}\big)
\end{align}
where 
\begin{align}\label{eq:zf}
\tilde{\xv}&=(H_{\sr}^TH_{\sr})^{-1}H_{\sr}^T \yv_{\sr}.
\end{align}
The SNR for $x_i$ is equal to $SNR_i = \frac{1}{\frac{1}{2}\sigma^2[(H_{\sr}^T H_{\sr})^{-1}]_{ii}}$.
Thus, an equivalent channel model can be written as
\begin{align}\label{eq:sr-zf}
\tilde{\yv} &=H_{\sr}^{\zf\rt}\xv +\tilde{\zv}
\end{align}
where $\tilde{\zv} \sim\mathcal{N}\big(0,\frac{1}{2}\sigma^2 I\big)$ and  $H_{\sr}^{\zf\rt}= \diag\Big(\frac{1}{\sqrt{[(H_{\sr}^T H_{\sr})^{-1}]_{11}}},\dots,\frac{1}{\sqrt{[(H_{\sr}^T H_{\sr})^{-1}]_{2N,2N}}}\Big) $.
Substituting $H_{\sr}^{\zf\rt}$ into \eqref{eq:pepR} instead of $H_{\sr}$, we obtain the PEP for the ZFaR.
Thus, we have $\tilde{H}_{\sr}=H_{\sr}^{\zf\rt}$ for the detection algorithms of the NML, DNwSRC, and ZGwSRC  in Section \ref{sec:wcsi}, and the SDRwSRC in Section \ref{sec:sdr}.
Since  $\frac{2}{\sqrt{[(H_{\sr}^T H_{\sr})^{-1}]_{ii}}}\sim \chi^2_{2(N_{\rt}-N+1)}$, the chi-squared distribution with $2(N_{\rt}-N+1)$ degree of freedom, the diversity order of the ZFaR is  $d_{\rt}=N_{\rt}-N+1$ \cite{Jiang--Varanasi--L12011}.
Subsequently, $d_{\rt}=N_{\rt}-N+1$ is used to determine the average error probability, $P_e$, in the detection algorithms of the NMLw2PEP,  DNwREP, and ZGwREP in Section \ref{sec:wscsi}, and in the SDRwREP detector in Section \ref{sec:sdr}. 

\item[3)] The SDR detector at the relay (SDRaR)

The SDRaR \cite{Jalden--Ottersten2008}, \cite{Luo--Ma--So--Ye--Zhang2010} with polynomial complexity is written as
\begin{align}
	\hat{\xv}=\sign([\Xv]_{1:2N,2N+1}) 
\end{align}
where $\Xv$ is the solution of the SDR optimization problem 
\begin{align}\label{eq:mimo_opt}
	&\text{ minimize } \tr(\Lv\Xv)\notag\\
	&\text{ subject to } [\Xv]_{ii}=1, i=1,\dots,2N+1 \notag\\
	&\qquad\qquad~~  \Xv\succeq 0,
\end{align}
and 
\begin{align*}
	&\Lv =
\begin{bmatrix}  
	H_{\sr}^TH_{\sr} & -H_{\sr}^T{\yv}_{\sr}\\
	-H_{\sr}^T{\yv}_{\sr} & {\yv}_{\sr}^T\,{\yv}_{\sr}
\end{bmatrix}.
\end{align*}
Even though the SDR problem in \eqref{eq:mimo_opt} relaxes the constraint of $\Xv=\begin{bmatrix}  
	\hat{{\xv}}\\
	1
\end{bmatrix}
\begin{bmatrix}  
	\hat{{\xv}}\\
	1
\end{bmatrix}^T$ to $\Xv\succeq 0$,
the SDR detection achieves the same diversity order as the ML detector \cite{Jalden--Ottersten2008}.
Thus, we use the original parameter ${H}_{\sr}$ for the SDRaR in the detection algorithms in Sections \ref{sec:wcsi}, \ref{sec:wscsi}, and \ref{sec:sdr}.

\item[4)] The detection network at the relay (DNaR)

  A MIMO detection network \cite{Samuel--Diskin--Wiesel2017} contains $L$ estimation layers, and each single layer is similar to Fig. \ref{fig:DFR_est_net} except that the part related to the helping signal $\hat{\xv}_{\rt,k}$ does not exist.
The input vector in the $k$th layer is 
\begin{align}\label{eq:mimo_input}
	\iv_k &=\begin{bmatrix} H_{\sr}^T\yv_{\sr}\\ \vv_{k-1}\\ \hat{\xv}_{k-1}\\ H_{\sr}^TH_{\sr}\hat{\xv}_{k-1}\end{bmatrix} 
\end{align}
and the parameters  
$${\bf \theta}=\{\Wv_{1k}, \bv_{1k},\Wv_{2k},\bv_{2k},\Wv_{3k},\bv_{3k}, \tv_{k}, k=1,\dots, L\} $$
are trained to minimize the loss function 
\begin{align}\label{eq:loss}
	l(\xv;\hat{\xv}_{\theta})=\sum_{k=1}^L \log(k+1)\frac{\|\xv-\hat{\xv}_k\|^2}{\|\xv-\tilde{\xv}\|^2}
\end{align}
where $\tilde{\xv}=(H_{\sr}^TH_{\sr})^{-1}H_{\sr}^T\yv_{\sr}$ is the ZF receiver in \eqref{eq:zf}.

As shown in \cite{Samuel--Diskin--Wiesel2017} and in Section \ref{sec:sim}, the MIMO detection network shows the similar performance with the SDR detector; thus,  we use the original parameter $H_{\sr}$ for the DNaR case.
\end{itemize}
\subsection{Detection Complexity}\label{sec:det_complexity}

For the complexity measure of the detection algorithms, we apply the naive calculation method, i.e., the complexity is $O(nmp)$ for the multiplication of matrices of $n\times m$ and $m\times p$, and $O(n^3)$ for the $n\times n$ matrix inversion.
The detection complexities for the DetR and DetD are discussed for constants $\beta_{\rt}=\frac{N_{\rt}}{N}$ and  $\beta_{\dt}=\frac{N_{\dt}}{N}$.

Based on the rules above regarding the computational complexity, the following results are obtained for the detection algorithms of the NML, NMLZG, and NMLDNs.
\begin{itemize}
\item Because $\big\|\yv_{\sd}-H_{\sd} \xv\big\|^2$, $\big\|\yv_{\rd}-H_{\rd} \xv_{\rt}\big\|^2$, and $\|H_{\sr}(\xv-\xv_{\rt})\|^2$ require the complexity of  $O(N^2)$ for each $\underline{\xv}=\begin{bmatrix}\xv^T&\xv_{\rt}^T\end{bmatrix}^T\in\mathcal{A}^{4N}$, the complexities of the NML detectors are at least $O(N^2 \cdot |\mathcal{A}|^{4N})$.
\item Due to the matrix multiplication and the matrix inversion, the NMLZG detectors require the complexity of $O(N^3)$.
For a quasi-static fading channel (fixed CSI), the NMLZG detectors only need to do a multiplication of a $4N\times 4N_{\dt}$ matrix,  $(H_{\dt}^TH_{\dt}+H_{\rt}^TH_{\rt})^{-1}H_{\dt}^T$, and a $4N_{\dt} \times 1$ vector, $\underline{\yv}$, and thus, the complexity is  $O(N^2)$.
\item For the NMLDNs,  the computation of the channel information $H_{\sd}^TH_{\sd}$, $H_{\rd}^TH_{\rd}$, and ${H}_{\sr}^T{H}_{\sr}$ requires the complexity of $O(N^{3})$, and the pre-computed values of $H_{\sd}^T\yv_{\sd}$ and $H_{\rd}^T\yv_{\rd}$ require the complexity of $O(N^{2})$.
The multiplication of $\iv_k$ and $\Wv_{1k}$ and the multiplications of $\uv_k$ and $\Wv_{2k}$, $\Wv_{\rt,2k}$, $\Wv_{3k}$ in each NNL require the complexity of $O(N^2)$; thus, the overall detection complexity for the NMLDN is  $O(N^{3})+O(LN^2)$.
For the quasi-static fading channels, the channel information $H_{\sd}^TH_{\sd}$, $H_{\rd}^TH_{\rd}$, and ${H}_{\sr}^T{H}_{\sr}$ does not need to be computed again; therefore, the complexity of  $O(LN^2)$ is obtained. 
Setting the number of layers as a multiple of $N$, the complexity becomes $O(N^{3})$.
The detection complexity of the NMLDNs can be lowered by reducing $L$ appropriately according to the required BER. 
\end{itemize}
The NMLSDR detectors possess the complexity of $O((4N)^{3.5}\log(1/\epsilon))=O(N^{3.5}\log(1/\epsilon))$ given a solution accuracy $\epsilon>0$ from \cite{Luo--Ma--So--Ye--Zhang2010}, \cite{Helmberg--Rendl--Vanderbei--Wolkowicz1996}.  
We summarize the detection complexity for the DetD as 
\begin{itemize}
\item NML: $O(N^2 \cdot |\mathcal{A}|^{4N})$  
\item NMLZG: $O(N^{2})$
\item NMLDN: $O(N^{3})$
\item NMLSDR: $O(N^{3.5}\log(1/\epsilon))$. 
\end{itemize}
Similarly, the detection complexities for the DetR mentioned in Section \ref{sec:DetR} are given as
\begin{itemize}
\item MLaR: $O(N^2 \cdot |\mathcal{A}|^{2N})$  
\item ZFaR: $O(N^{2})$
\item DNaR: $O(N^{3})$
\item SDRaR: $O(N^{3.5}\log(1/\epsilon))$.
\end{itemize}

\subsection{System  Configuration}\label{sec:system_config}

Using various DetR and DetD, the DF relay system in Fig. \ref{fig:DF} exhibits different characteristics in error performance and detection complexity.
We introduce and compare  several types of system methods depending on the applied detection algorithms at the relay and the destination.
The complexity is based on the discussion in Section \ref{sec:det_complexity}, and the error performance will be demonstrated in Section \ref{sec:sim}.

\begin{itemize}
\item MLaR-NML: 

Exhaustive search detection algorithms are implemented at the relay and the destination, i.e., the optimal ML is used at the relay, and the suboptimal NML detectors including the NML, NMLw2PEP, and NMLwRSD are applied at the destination. 
This system achieves excellent performance, but cannot be used in large-scale antenna systems due to its high complexity.

\item ZFaR-NMLZG: 

A ZF detector is used at the relay, and NMLZG detectors such as the ZGwSRC, ZGwREP, and ZGwRSD are applied at the destination. 
This method has a simple detection complexity at both the relay and the destination, but exhibits poor performance.

\item SDRaR-NMLSDR: 

The SDR versions of the exhaustive search detectors are implemented at the relay and the destination. 
The NMLSDR detectors include the SDRwSRC, SDRwREP, and SDRwRSD detectors in Section \ref{sec:sdr}.
This method exhibits a fair performance with polynomial complexity.

\item DNaR-NMLDN: 

The deep learning detection networks are employed at both the relay and the destination. 
The applied NMLDN methods include the DNwSRC, DNwREP,  DNwRSD, and sDNwRSD proposed in Section \ref{sec:DNaD}.
This method achieves a reasonable performance with lower complexity through a pre-process of training.

\end{itemize}

\section{Performance Evaluation}\label{sec:sim}

\begin{figure}[t]
\centering{\includegraphics[width=3.4in]{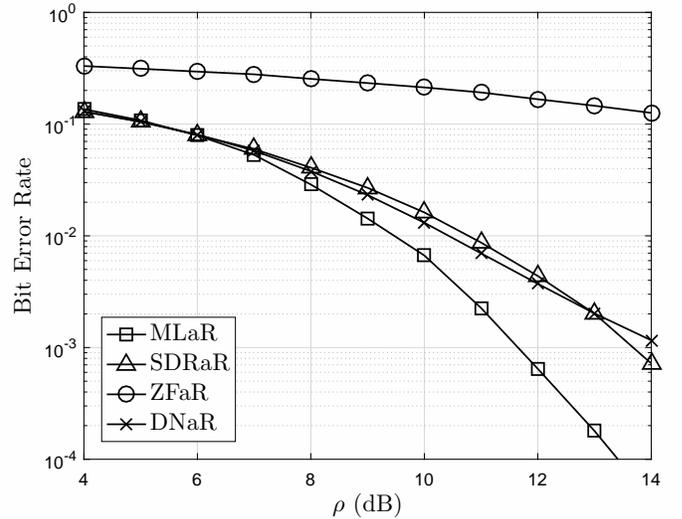}}
\caption{BER comparisons of various DetR over the SR channel with $N_{\rt}=N=10$. 
}\label{fig:DetaR}
\end{figure}
\begin{figure}[h]
\centering{\includegraphics[width=3.4in]{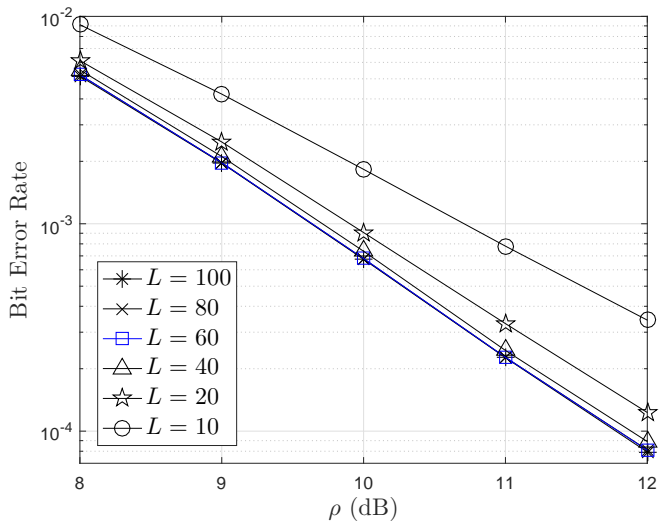}}
\caption{BER comparisons of the DNaR-DNwSRC method trained by the loss function $l_1$ in \eqref{eq:l1} for various numbers of layers over the MIMO DF relay channel with $N_{\rt}=N_{\dt}=N=10$. 
}\label{fig:d_dec_layers}
\end{figure}
\begin{figure}[h]
\centering{\includegraphics[width=3.4in]{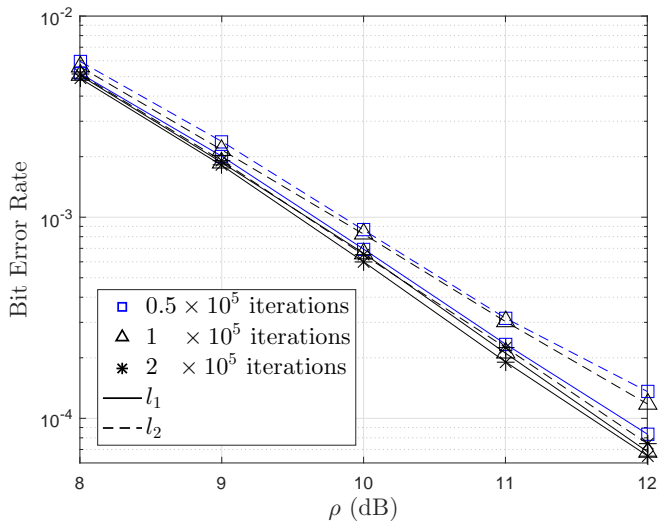}}
\caption{BER comparisons of the DNaR-DNwSRC method for various training iterations over the MIMO DF relay channel with $N_{\rt}=N_{\dt}=N=10$.  
}\label{fig:d_dec_train}
\end{figure}
\begin{figure}[t]
	\subfigure[]{\includegraphics[width =
	3.2in]{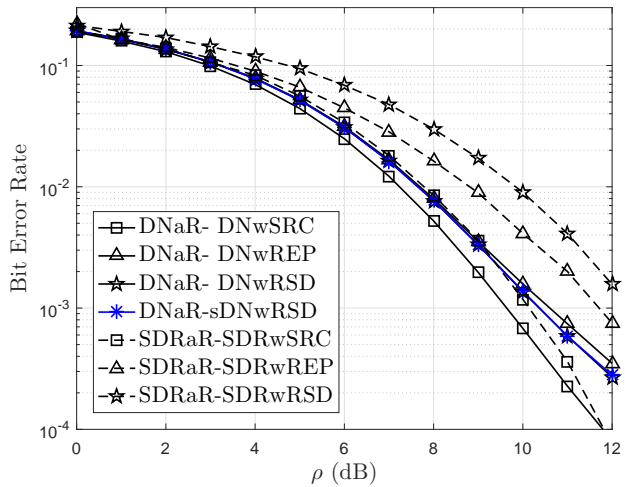}\label{fig:d_dec_l1}} ~~
	\subfigure[]{\includegraphics[width
	=3.2in]{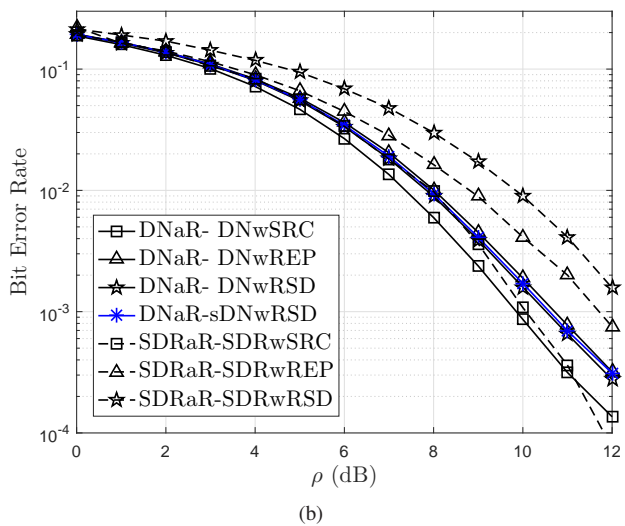}\label{fig:d_dec_l2}}
\caption {BER comparisons of the DNaR-NMLDN and SDRaR-NMLSDR methods over the MIMO DF relay channel with $N_{\rt}=N_{\dt}=N=10$.
(a) The NMLDNs are trained using $l_1$; (b) The NMLDNs are trained using $l_2$. }\label{fig:d_dec}
\end{figure}
\begin{figure}[h]
\centering{\includegraphics[width=3.4in]{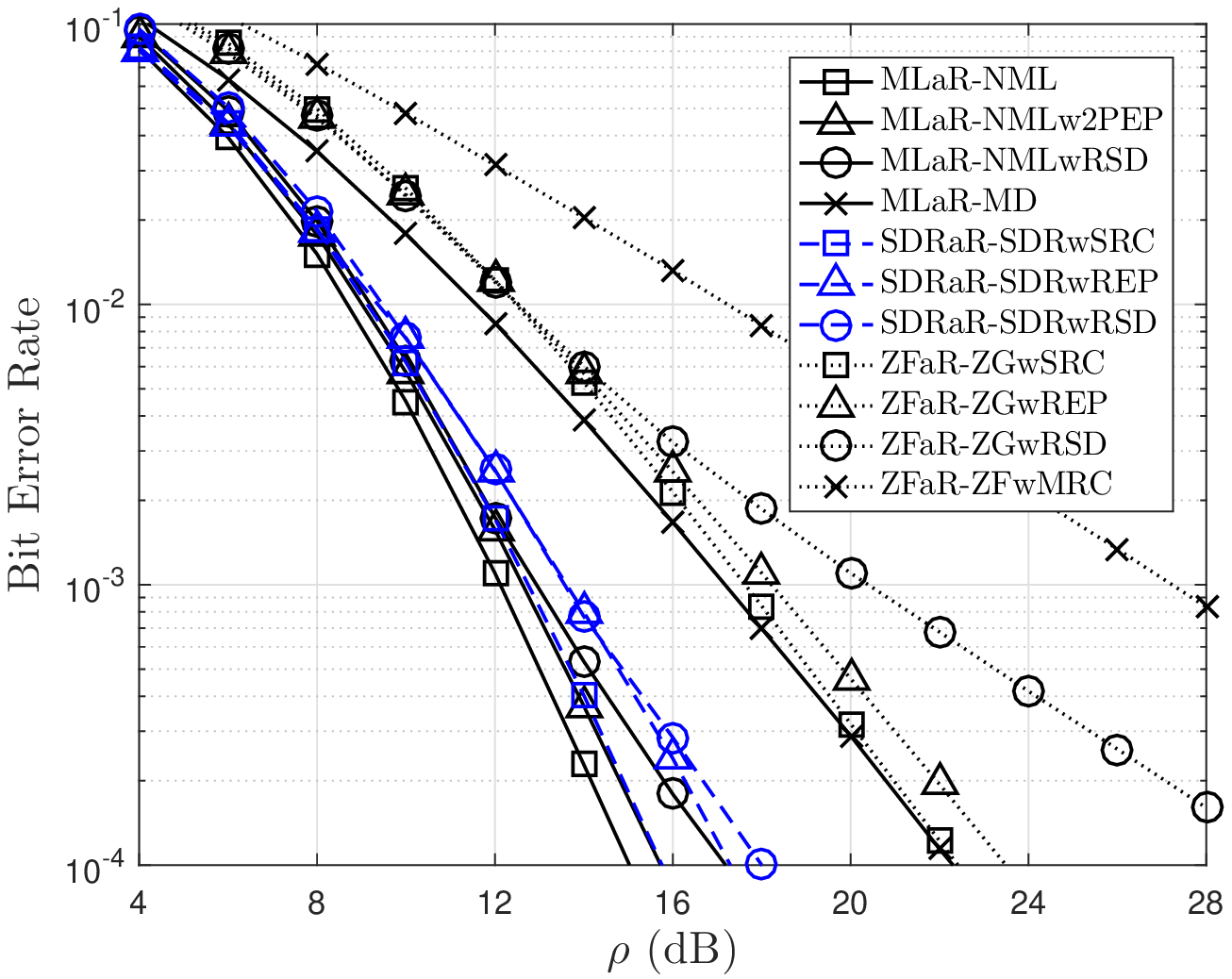}}
\caption{BER comparisons of various DetR and DetD methods over the MIMO DF relay channel with $N_{\rt}=N_{\dt}=N=2$.
}\label{fig:MIMO2_2detaD}
\end{figure}

In  this section, we evaluate the performance of various system methods and compare the proposed detection algorithms.
We consider  MIMO  DF relay channel with $H_{\sr}^C\sim\CN(0,I_{N_{\rt}N})$, $H_{\sd}^C\sim\CN(0,I_{N_{\dt}N})$, and $H_{\rd}^C\sim\CN(0,I_{N_{\dt}N})$ in \eqref{eq:srC}-\eqref{eq:rdC}.

First of all, we compare the error performance of the DetR mentioned in Section \ref{sec:DetR} for $N_{\rt}=N=10$.
As shown in Fig. \ref{fig:DetaR}, the MLaR achieves the best performance, and the ZFaR exhibits the poorest performance.
The DNaR achieves the error performance similar to the SDRaR over a fairly wide SNR range.
This supports the argument in Session \ref{sec:DetR} well.
Note that the DNaR in Fig. \ref{fig:DetaR} has undergone the structure in Session \ref{sec:DetR}\,-\,4) and has been trained by $N_{\text{iteration}}=10^6$ iterations with $N_{\text{batch}}=10^4$ batches in each iteration.

Since the MLaR-NML methods could not be implemented in real time due to their high complexity, we compare the DNaR-NMLDN and SDRaR-NMLSDR methods for $N_{\dt}=N_{\rt}=N=10$.
The similar performance of the DNaR and the SDRaR shown in Fig. \ref{fig:DetaR} as well as in \cite{Samuel--Diskin--Wiesel2017} renders a fair comparison.
For the best-performance NML detectors and the poorest-performance NMLZG detectors, we evaluate later with a smaller number of antennas.
We first evaluate the BERs of the DNaR-DNwSRC method with various numbers of layers trained  $N_{\text{iteration}}=5\times 10^4$ iterations as shown in Fig. \ref{fig:d_dec_layers}.
As the number of layers increases, the BER improves and converges to a certain level when $L\ge 60=3*2N $.
Hence, $L=6N$ is applied in the following simulations.
On the other hand, the BERs of the DNaR-DNwSRC method with different numbers of training iterations are shown in Fig. \ref{fig:d_dec_train}.
It can be seen that as the number of the training iterations increases, the error performance improves. 
It is possible to obtain better performance by increasing the number of training iterations, but considering the time required for training, we train the NMLDNs with $N_{\text{iteration}}=5\times 10^4$ iterations when comparing with the NMLSDR methods.
Fig. \ref{fig:d_dec} compares BERs for various DNaR-NMLDN and SDRaR-NMLSDR  methods over the  relay channel.
From the curves, one can observe as the following:
\begin{itemize}
\item The DNaR-NMLDN methods trained using the loss function $l_1$ achieve better performance than those trained by the loss function $l_2$ and the SDRaR-NMLSDR methods, under the same knowledge of the SR link in wide ranges of SNR.

\item The DNwSRC achieves the best error performance, and the DNwREP,  DNwRSD, and sDNwRSD show similar performance for the  DNaR.

\item Without knowledge of the instantaneous CSI of the SR link, the DNaR-DNwREP, DNaR-DNwRSD, and sDNaR-DNwRSD methods achieve better performance than the SDRaR-SDRwREP and SDRaR-SDRwRSD methods. 
Particularly, without any knowledge of the SR link at the destination, the DNaR-DNwRSD and DNaR-sDNwRSD show significant SNR improvements compared to the SDRaR-SDRwRSD, i.e., approximately 2 dB at BER$=10^{-2}$. 
\end{itemize}

Additionally, we evaluate the MLaR-NML, SDRaR-NMLSDR, and ZFaR-NMLZG methods in the DF relay channel with $N_{\dt}=N_{\rt}=N=2$. 
The DNaR-NMLDNs are not compared in this case since the deep learning detection does not have any advantages in both performance and complexity in small antenna systems.
Fig. \ref{fig:MIMO2_2detaD} shows that the MLaR-NML methods obtain the best performance, and the ZFaR-NMLZG methods exhibit the worst performance, while the SDRaR-NMLSDR methods show a nice performance with slopes similar to the corresponding MLaR-NML methods.
Moreover, the NMLwRSD, ZGwSRC, ZGwREP, and ZGwRSD detectors proposed in Propositions \ref{prop:nmlwrsd}, \ref{prop:ldwsrc}, \ref{prop:ldwrep}, \ref{prop:ldwrsd} exhibit good performance compared to the existing detectors under the same DetR. 
In detail, the MLaR-NMLwRSD method obtains approximately $3.6$ dB SNR improvement compared with the MLaR-MD method at BER$=10^{-3}$ without any knowledge of the SR link.
The ZGwSRC, ZGwREP, and ZGwRSD detectors yield approximately $9.6$ dB, $9$ dB, and  $6.8$ dB SNR improvements, respectively, compared to the ZFwMRC detector at BER$=10^{-3}$ when the ZF detector is used at the relay.

\section{Conclusions}

Based on the exhaustive search suboptimal NML detectors, the deep learning detection networks  are proposed by unfolding iterative calculations into neural-network layers in the MIMO DF relay channel with the instantaneous or statistical CSI of the SR link.
Without any knowledge of the SR channel, the suboptimal NMLwRSD detector is proposed by reflecting the influence of the SR channel using the squared relay signal distance.
Based on the NMLwRSD detector, two deep learning detection networks, the DNwRSD and sDNwRSD, are also proposed that do not take into account the CSI of the SR link. 
The deep learning detection networks exhibit a fair performance with less complexity compared to the suboptimal NML detectors and the NMLSDR detectors.
The proposed detection algorithms of the NMLwRSD, SDRwSRC, SDRwREP, SDRwRSD, ZGwSRC, ZGwREP, ZGwRSD, DNwSRC, DNwREP, DNwRSD, and sDNwRSD can be applied  in more complex communication networks such as multi-relay channels and multi-way relay channels.
Furthermore,  the discussion on the performance and complexity for various types of DetR-DetD methods provides a basic idea and direction for the system configuration.


%

\begin{thebibliography}{99}

\bibitem{van-der-Meulen1971b}
E.~C. van~der Meulen, ``The discrete memoryless channel with two
  senders and one receiver,'' in \emph{Proc. 2nd Int. Symp. Inf. Theory},
  Tsahkadsor, Armenian SSR, 1971, pp. 103--135.

\bibitem{Cover--El-Gamal1979} T.~M. Cover and A.~El~Gamal, ``Capacity theorems for the relay channel,''   \emph{{IEEE} Trans. Inf. Theory}, vol.~25, no.~5, pp. 572--584, Sep. 1979.


\bibitem{Laneman--Tse--Wornell2004} J. N. Laneman, D. N. C. Tse and G. W. Wornell, ``Cooperative diversity in wireless networks: Efficient protocols and outage behavior,'' {\em IEEE Trans. Inf. Theory}, vol. 50, no. 12, pp. 3062--3080, Dec. 2004.

\bibitem{Sendonaris2}A. Sendonaris, E. Erkip, and B. Aazhang, ``User cooperation diversity-
Part II: Implementation aspects and performance analysis,'' {\em
IEEE Trans. Commun.}, vol.~51, no. 11, pp.~1939--1948, Nov.~2003.

\bibitem{Wang--Cano--Giannaki--Laneman2007} T. Wang, A. Cano, G. B. Giannakis, and J. N. Laneman, ``High performance cooperative demodulation with decode-and-forward relays,''
{\em IEEE Trans. Commun.}, vol. 55, no. 7, pp. 1427--1438, Jul. 2007.


\bibitem{Jin2011ieice} X. Jin, D.-S. Jin, J.-S. No, and D.-J. Shin,
``Diversity analysis of MIMO decode-and-forward relay network by using near-ML decoder,'' 
{\em IEICE Trans. Commun.}, vol. E94-B, no. 10, pp. - Oct. 2011.

\bibitem{Jin2011twc} X. Jin, J.-S. No, and D.-J. Shin,
``Relay selection for decode-and-forward cooperative network with
multiple antennas,'' {\em IEEE Trans. Wireless Commun.}, vol. 10, no 12, pp. 4068--4079,  Dec. 2011.

\bibitem{Jin2014jcn} X. Jin, E. J. Kum and D. W. Lim, ``Maximum diversity achieving decoders in MIMO decode-and-forward relay systems with partial CSI,'' {\em J. Commun. Networks}, vol. 16, no. 1, pp. 26--35, Feb. 2014.



\bibitem{Chalise--Vandendorpe2009} B. K. Chalise and L. Vandendorpe, ``Performance analysis of linear receivers in a MIMO relaying system,'' {\em IEEE Commun. Letters}, vol. 13, no. 5, pp. 330--332, May 2009.

\bibitem{Jin--Kim2017} X.~Jin and Y.~Kim, ``The approximate capacity of the {MIMO} relay channel,''
  \emph{{IEEE} Trans. Inf. Theory}, vol.~63, no.~2, pp. 1167--1176, Feb 2017.

\bibitem{Lupas--Verdu1989}
R.~Lupas and S.~Verdu, ``Linear multiuser detectors for synchronous  code-division multiple-access channels,'' \emph{{IEEE} Trans. Inf. Theory},  vol.~35, no.~1, pp. 123--136, Jan 1989.


\bibitem{Goodfellow--Bengio--Courville2016} I. Goodfellow, Y. Bengio, and A. Courville, {\it Deep Learning}. MIT Press,
Nov. 2016.


\bibitem{Ioannidou--Chatzilari--Nikolopoulos--Kompatsiaris} A. Ioannidou, E. Chatzilari, S. Nikolopoulos, and I. Kompatsiaris, ``Deep learning advances in computer vision with 3d data: A survey,'' {\em ACM Computing Surveys (CSUR)}, 50, no. 2,  2017.

\bibitem{Abdel-Hamid--Mohamed--Jiang--Deng--Penn--Yu2014} O. Abdel-Hamid, A. r. Mohamed, H. Jiang, L. Deng, G. Penn and D. Yu, ``Convolutional neural networks for speech recognition,'' {\em IEEE/ACM Transactions on Audio, Speech, and Language Processing}, vol. 22, no. 10, pp. 1533--1545, Oct. 2014.

\bibitem{Zhang--Patras--Haddadi2018} C. Zhang, P. Patras, and H. Haddadi,
``Deep learning in mobile and wireless networking: A survey,'' {\em arXiv preprint arXiv:1803.04311}, 2018.


\bibitem{Nachmani--Marciano--Lugosch--Gross--Burshtein--Beery} E. Nachmani, E. Marciano, L. Lugosch, W. J. Gross, D. Burshtein and Y. Be’ery, ``Deep learning methods for improved decoding of linear codes,'' {\em  IEEE Journal of Selected Topics in Signal Processing,} vol. 12, no. 1, pp. 119--131, Feb. 2018.
\bibitem{Liang--Shen--Wu2018} F. Liang, C. Shen, and F. Wu, ``An iterative BP-CNN architecture for channel decoding,'' {\em IEEE Journal of Selected Topics in Signal Processing}, vol. 12, no. 1, pp. 144--159, Feb. 2018.

\bibitem{Yan--Long--Wang--Fu--Ou--Liu2017} X. Yan, F. Long, J. Wang, N. Fu, W. Ou, and B. Liu, ``Signal detection of MIMO-OFDM system based on auto encoder and extreme learning machine,'' in {\em International Joint Conference on Neural Networks (IJCNN)},  Anchorage, AK, May 14-19, 2017,  pp. 1602-–1606.

\bibitem{Samuel--Diskin--Wiesel2017} N. Samuel, T. Diskin and A. Wiesel, ``Deep MIMO detection,'' in {\em IEEE 18th International Workshop on Signal Processing Advances in Wireless Communications (SPAWC)}, Sapporo, Jul. 3-6, 2017, pp. 1--5.
\bibitem{Farsad--Goldsmith2017}  N. Farsad and A. Goldsmith,``Detection algorithms for communication systems using deep learning,'' {\em arXiv preprint arXiv:1705.08044}, 2017.

\bibitem{Farsad--Goldsmith2018}  N. Farsad and A. Goldsmith,``Neural network detection of data sequences in communication systems,'' {\em arXiv preprint arXiv:1802.02046}, 2018.


\bibitem{Hershey--Roux--Weninger2014} J. R Hershey, J. L. Roux, and F. Weninger, ``Deep unfolding: Model-based inspiration of novel deep architectures,'' {\em arXiv preprint arXiv:1409.2574}, 2014.

\bibitem{MIMO_boundary}H. Art$\acute{e}$s, D. Seethaler, and F. Hlawatsch, ``Effcient detection algorithms
for MIMO channels: A geometrical approach to approximate ML detection,''
{\em IEEE Trans. Signal Processing}, vol. 51, no. 11, pp. 2808--2820, Nov. 2003.

\bibitem{Hochwald--Brink2003} B. M. Hochwald and S. T. Brink, ``Achieving near-capacity on a multiple-antenna channel,'' {\em IEEE Trans. Commun.}, vol. 51, pp. 389–-399, Mar. 2003.
\bibitem{ECC_FA}S. Lin and D. J. Costello, Jr., {\em Error Control Coding: Fundamentals and
Applications}, 2nd ed. Upper Saddle River, NJ: Pearson Prentice Hall, 2004.


\bibitem{Ju--Kim2009} M. C. Ju and I. M. Kim, ``ML performance analysis of the decode-and-forward protocol in cooperative diversity networks,'' {\em IEEE Trans. Wireless Commun.}, vol. 8, no. 7, pp. 3855–-3867, Jul. 2009.


\bibitem{He--Zhang--Ren--Sun2016}K. He, X. Zhang, S. Ren and J. Sun, ``Deep residual learning for image recognition,'' 2016 IEEE Conference on Computer Vision and Pattern Recognition (CVPR), Las Vegas, NV, 2016, pp. 770--778.

\bibitem{Boyd--Vandenberghe2004}
S.~Boyd and L.~Vandenberghe, \emph{Convex Optimization}.\hskip 1em plus 0.5em
  minus 0.4em\relax Cambridge: Cambridge University Press, 2004.

\bibitem{tensorflow}M. Abadi, A. Agarwal, P. Barham, E. Brevdo, Z. Chen, C. Citro, G. S. Corrado, A. Davis, J. Dean, M. Devin, et al., 
``Tensorflow: Large-scale machine learning on heterogeneous distributed systems,'' arXiv preprint,
2016. url  = {http://arxiv.org/abs/1603.04467}



\bibitem{Kingma--Ba2014} D. Kingma and J. Ba, ``Adam: A method for stochastic optimization,'' arXiv preprint arXiv:1412.6980, 2014.

\bibitem{Jalden--Ottersten2008} J. Jald$\acute{e}$n and B. Ottersten, ``The diversity order of the semidefinite relaxation detector,'' {\em IEEE Trans. Inf. Theory}, vol. 54, no. 4, pp. 1406--1422, 2008.
\bibitem{Luo--Ma--So--Ye--Zhang2010} Z. Q. Luo, W. K. Ma, A. M. So, Y. Ye, and S. Zhang, ``Semidefinite relaxation of quadratic optimization problems,'' {\em IEEE Signal Processing Magazine}, vol. 27, no. 3, pp. 20–-34, 2010.

\bibitem{Grant--Boyd2013} M.~Grant and S.~Boyd, ``{CVX}: Matlab software for disciplined convex programming, version 2.1 beta,''  Dec. 2017.


\bibitem{Jiang--Varanasi--L12011}Y. Jiang, M. K. Varanasi and J. Li, ``Performance analysis of ZF and MMSE equalizers for MIMO systems: An in-depth study of the high SNR regime,'' {\em IEEE Trans. Inf. Theory}, vol. 57, no. 4, pp. 2008--2026, April 2011.

\bibitem{Helmberg--Rendl--Vanderbei--Wolkowicz1996} C. Helmberg, F. Rendl, R. Vanderbei, and H. Wolkowicz,  ``An interior-point method for semidefinite programming,'' {\em SIAM J. Optim.}, vol. 6, pp. 342–-361, 1996.

\end{thebibliography}
\end{document}